%% file: main.tex
\documentclass{article}

\usepackage{microtype}
\usepackage{graphicx}
\usepackage{subfigure}
\usepackage{booktabs} 

\usepackage{hyperref}

\usepackage[accepted]{icml2024}

\usepackage{amsmath}
\usepackage{amssymb}
\usepackage{mathtools}
\usepackage{amsthm}

\usepackage[capitalize,noabbrev]{cleveref}

\theoremstyle{plain}

\theoremstyle{definition}

\theoremstyle{remark}

\usepackage[textsize=tiny]{todonotes}
\usepackage{multirow,placeins,makecell,dsfont}

\icmltitlerunning{Tabular Data Class-Conditioned Augmentation}

\begin{document}

\twocolumn[
\icmltitle{Tabular Data Contrastive Learning via Class-Conditioned and Feature-Correlation Based Augmentation}



\icmlsetsymbol{equal}{*}

\begin{icmlauthorlist}
\icmlauthor{Wei Cui}{lyr6}
\icmlauthor{Rasa Hosseinzadeh}{lyr6}
\icmlauthor{Junwei Ma}{lyr6}
\icmlauthor{Tongzi Wu}{lyr6}
\icmlauthor{Yi Sui}{lyr6}
\icmlauthor{Keyvan Golestan}{lyr6}
\end{icmlauthorlist}

\icmlaffiliation{lyr6}{Layer 6 AI, Toronto, Canada}

\icmlcorrespondingauthor{Wei Cui}{wei@layer6.ai}

\icmlkeywords{Tabular Data, Contrastive Learning, Self-Supervised Learning, Data Augmentation}

\vskip 0.3in
]



\printAffiliationsAndNotice{}  

\begin{abstract}
Contrastive learning is a model pre-training technique by first creating similar views of the original data, and then encouraging the data and its corresponding views to be close in the embedding space. Contrastive learning has witnessed success in image and natural language data, thanks to the domain-specific augmentation techniques that are both intuitive and effective. Nonetheless, in tabular domain, the predominant augmentation technique for creating views is through corrupting tabular entries via swapping values, which is not as sound or effective. We propose a simple yet powerful improvement to this augmentation technique: corrupting tabular data conditioned on class identity. Specifically, when corrupting a specific tabular entry from an anchor row, instead of randomly sampling a value in the same feature column from the entire table uniformly, we only sample from rows that are identified to be within the same class as the anchor row. We assume the semi-supervised learning setting, and adopt the pseudo labeling technique for obtaining class identities over all table rows. We also explore the novel idea of selecting features to be corrupted based on feature correlation structures. Extensive experiments show that the proposed approach consistently outperforms the conventional corruption method for tabular data classification tasks. Our code is available at \url{https://github.com/layer6ai-labs/Tabular-Class-Conditioned-SSL}.
\end{abstract}

\input{introduction}

\input{problem}

\input{method}

\input{experiments}

\section{Conclusion}
\label{sec:conclusion}
We explored different variations in tabular data corruption based augmentation approach used in contrastive learning, under the semi-supervised learning setting. Specifically, to ensure high similarities between the augmented views and the anchor row, we proposed both the class-conditioned corruption approach and the correlation-based feature masking approach. Compared to the conventional corruption approach, we aim to incorporate more information into the corruption procedure, specifically, the class information obtained via pseudo-labeling; and the feature correlation information obtained via trained XGBoost models. Extensive experimental results suggest that the class-conditioned corruption approach supports learned embedding spaces that better segment between classes, and therefore provides improvements to down-stream classification accuracy. On the other hand, the correlation-based feature masking approach fails to show consistent improvements compared to random selection of corrupted features. We hypothesize this is largely due to the fact the preprocessing stages in obtaining the benchmark datasets have removed much redundancy, resulting sets of features sharing little to none correlation within each other. We hope this paper sheds light on further explorations and improvements to the challenging problem of effective tabular data augmentation.

\section*{Acknowledgements}
We sincerely thank our colleague Gabriel Loaiza-Ganem for the valuable discussion and ideas, which sparked one of the exploration directions in this paper on the selective subset of features for corruption.

\bibliography{citations}
\bibliographystyle{icml2024}

\newpage
\input{appendices}

\end{document}

%% file: introduction.tex
\section{Introduction}
\label{sec:intro}

Tabular data is one of the most fundamental and prevalent data modalities, from which rich information and insights can be extracted through proper learning algorithms. Nonetheless, the deep learning algorithm, arguably the most widely used learning algorithms, currently still struggles to excel in tabular domain tasks \citet{grinsztajn22}, which stresses the need for algorithm improvement. On the other hand, just like in the other data modalities, with the sheer volume of tabular data being present, it is often too time or resource consuming to collect quality target label for each data point. Correspondingly, semi-supervised learning \citet{semisupervised} or self-supervised learning techniques \citet{selfsupervised} are required for model pre-training to obtain high-performance models on tabular data.

In recent years, \emph{contrastive learning} \citet{simclr} has emerged as a highly popular pre-training technique frequently adapted in the self-supervised learning and semi-supervised learning settings. Contrastive learning is mainly used to pre-train an encoder block, which maps the input raw data to an intermediate embedding space. It has been successfully utilized in various applications with different data modalities \citet{simclr,simsiam,shuangma21,videomoco}. To perform contrastive learning, as the first step, researchers engineer data augmentation techniques to create similar \emph{views} on top of each original data point (often referred to as the \emph{anchor}). A contrastive loss is then optimized, which minimizes a distance metric (\textit{i.e.,} cosine similarity) in the embedding space between the anchor and its corresponding views. In the classical contrastive loss formulation, views generated from different anchors are explicitly pushed away in the embedding space. However, multiple variants to the contrastive loss have been proposed that either do not explicitly push away dissimilar pairs \citet{simsiam}, or incorporate additional terms into the loss function \citet{vicreg}.

Naturally, the design of the data augmentation techniques is pivotal for the contrastive pre-training performance. A good data augmentation technique often incorporates domain-specific knowledge. For image data, researchers have successfully explored various augmentation techniques that lead to image views with rich varieties, while preserving the original semantic meanings, including rotation, cropping, color distortions, Gaussian blurring, and so on \citet{simclr,simsiam}. Similarly, in the domain of natural language processing (NLP), contrastive learning has also witnessed success through employing domain-specific augmentation techniques, including word masking, word replacement, reordering, and so on \citet{shorten21}. 

Nonetheless, for the domain of tabular data, it is more challenging to design such effective data augmentation techniques. The reasons are multifold: 
\begin{itemize}
    \item \textbf{Lack of Structure}: tabular data in general lacks spatial or temporal structures that can be exploited for augmentation techniques such as rotation or re-ordering.
    \item \textbf{Distinct Schemas}: the list of features in each table is unique in ordering and semantics. The organization and representation of information is highly distinct from table to table. 
    \item \textbf{Heterogeneous Features}: with tabular features of different types, units, or ranges, an augmentation technique applicable to some features might be inapplicable to other features within the same table. 
\end{itemize}
There are three main classes of augmentation techniques for tabular data in the literature: 1. Feature-value Corruption \citet{vime,tabbie,scarf,stunt}, 2. Inter-sample Weighted Summation \citet{saint,contramix}, and 3. Cropping Subsets of Features \citet{subtab,transtab}. Among the three classes of approaches, the feature-value corruption is the most adopted technique. Regardless, due to the above-mentioned characteristics of tabular data, these augmentation techniques have yet to show significant improvements when being used for model pre-training.

In this paper, recognizing the gap in tabular data augmentation techniques, we first propose an easy yet powerful improvement to the feature-value corruption technique, by incorporating the class information into the corruption procedure. Specifically, when corrupting a feature value on the anchor row, instead of sampling uniformly across the entire table for the replacement, we restrict the sampling within rows that belong to the same class as the anchor row. We refer to this improvement as \emph{class-conditioned corruption}. Through this simple modification, we further ensure that the generated views are more likely to be semantically similar to the original anchor, and therefore should be pushed closer to the anchor in the embedding space via contrastive learning. As we do not have the target information for every row under the semi-supervised setting, we adopt the \emph{pseudo-labeling} technique to obtain estimations of targets. During the pre-training process, we iteratively train our encoder and update our targets. 

The class-conditioned corruption technique can be regarded as \emph{how to corrupt} when generating augmented tabular rows. We further explore into another dimension of corruption strategy: \emph{where to corrupt}. Specifically, to improve upon the current approach, we exploit feature correlations, and attempt to sample the subset of features based on the correlation structure, which we refer to as \emph{correlation-based feature masking}. We conducted extensive experiments on the OpenML-CC18 dataset \citet{openmlcc} for both avenues. Results suggest that the class-conditioned corruption improvement boosts the pre-trained model to outperform its counterpart, while the correlation-based feature masking technique shows no concrete evidence for improvement.

For mathematical symbols, we use the superscript $(\cdot)^\text{T}$ to denote the transpose of a matrix (or a vector regarded as a single-column matrix). We use $||\cdot||_2$ to denote the $L^2$ norm of a vector. We use $[N]$ to denote the integer set containing natural numbers up to $N$; and $\lceil \cdot \rceil$ to denote the ceiling integer function. Lastly, we use the operator $\leftarrow$ to denote the assignment to a specified variable.

%% file: problem.tex
\section{Problem Formulation \& Background}
\label{sec:prob}
\subsection{Semi-Supervised Tabular Data Classification}\label{sec:prob_I}
Consider a tabular dataset $D=(D_{\text{train}}, D_{\text{test}})$, where $D_{\text{train}}=\{(x_i, y_i)\}_{i=1}^{N_l}\cup\{x_i\}_{i=N_l+1}^{N_l+N_u}$ consists of $N_l$ labeled samples and $N_u$ unlabeled samples for training; and \mbox{$D_{\text{test}}=\{(x_i, y_i)\}_{i=N_l+N_u+1}^{N_l+N_u+N_t}$} consists of $N_t$ labeled samples for testing, with the testing labels $\{y_i\}_{i=N_l+N_u+1}^{N_l+N_u+N_t}$ unknown to the model. Each sample input $x_i$ consists of $M$ features, with each feature being either categorical or numerical in type. We use $x_i^{(k)}$ to denote the $k$-th feature of the $i$-th sample. For each input in the labeled set, a class label $y_i$ is available $\forall i\in[N_l]$. We aim to learn a parametrized mapping $f_\theta(\cdot)$ that solves the classification task by mapping each input in the testing set to its class.

We explore the use of a neural network for modeling $f_\theta$, with $\theta$ corresponding to the neural network parameters. Based on the contrastive learning algorithm, the neural network consists of three parts:
\begin{enumerate}
    \item An encoder, with parameters denoted as $\theta_e$.
    \item A pre-train head, with parameters denoted as
    $\theta_p$.
    \item A classification head,
    with parameters denoted as $\theta_c$.
\end{enumerate}
We then have $\theta=\{\theta_e, \theta_p, \theta_c\}$ as the model parameters space. Note that we adopt to this three-part architecture following the convention of contrastive learning research \citet{simclr}, where the pre-train head $\theta_p$ is only included in the pre-training stage to allow for extra flexibility in the learned representations. It will be discarded in the down-stream fine-tuning stage.

\subsection{Contrastive Learning}\label{sec:prob_II}
Contrastive learning is the process of pre-training the model to learn an embedding space which will be further used for down-stream classification or regression tasks. Below, we present the original contrastive learning approach, which is adapted by our approaches. We note that the contrastive learning variants later proposed are equally applicable to our approach and discussions. 

Let $g(\cdot)$ denote the data augmentation function. Given an anchor $x_i$ from the training set, a view is generated as follows:
\begin{align}
    \tilde{x}_i=g(x_i)\:.
\end{align}
With $g(\cdot)$ intended to be a semantic preserving operation, $\tilde{x}_i$ should be highly similar to $x_i$. Therefore, their corresponding representations in the embedding space should also be close.

After obtaining views as above, we compute the embeddings for the original data points as well as the views: 
\begin{align}
    z_i =& f_{\theta_p}(f_{\theta_e}(x_i)) \label{equ:emb_orig}\\
    \tilde{z}_i =& f_{\theta_p}(f_{\theta_e}(\tilde{x}_i)) \label{equ:emb_aug}.
\end{align}
Consider a sampled training batch $\{x_i\}_{i=1}^N$ with batch size $N$. Following \cref{equ:emb_orig} and \cref{equ:emb_aug}, we can obtain two sets of embeddings: $\{z_i\}_{i=1}^N$ and $\{\tilde{z}_i\}_{i=1}^N$. We concatenate the two sets of embeddings, leading to the set of $2N$ embeddings $\{\hat{z}_i\}_{i=1}^{2N}$, in which the first and second halves correspond to $\{z_i\}$ and $\{\tilde{z}_i\}$, respectively. We denote the cosine similarity between any pair of embeddings $z_i$ and $z_j$ as follows:
\begin{align}
    s_{i,j} =& \frac{\hat{z}_i^T\hat{z}_j}{||\hat{z}_i||_2||\hat{z}_j||_2} \:.
\end{align}

A contrastive loss is then defined on this training batch as follows: 
\begin{align}\label{equ:contrastive_loss}
    \mathcal{L} \doteq \frac{1}{2N}\sum_{i=1}^{2N} -\log\left(\frac{e^{s_{i,i'}/\tau}}{\sum_{j=1}^{2N}\mathds{1}_{[j\neq i]} e^{s_{i,j}/\tau}} \right) ,
\end{align}
where $i'\doteq (i+N)\mod{N}$ is the index of the pairing view (or anchor) embedding to the index $i$.
By minimizing \cref{equ:contrastive_loss}, we gradually push closer pairs of anchor and view, and push away anchors and views across different pairs to avoid collapsing to the trivial solution.

After optimizing the model by minimizing \cref{equ:contrastive_loss}, as mentioned in \cref{sec:prob_I}, we will proceed to discard the pre-train head $\theta_p$, freeze the encoder portion $\theta_e$, and train the classification head $\theta_c$ taking the outputs from the encoder as inputs on the labeled training set.

\subsection{Data Augmentation via Random Corruption}\label{sec:prob_III}

As described in \cref{sec:prob_II}, the data augmentation function $g(\cdot)$ serves as an important piece to the entire pre-training process. In the literature, the most popular augmentation technique for tabular data has been the feature value corruption. Randomly sampling has been used for both sampling the replacement value within each feature, as well as sampling which subset of features to be corrupted \citet{vime,tabbie,scarf,stunt}.

To describe the corruption procedure in details, let $p$ be the percentage of features to be corrupted, which is a hyper-parameter that can be chosen arbitrarily. We would then have $\lceil Mp \rceil$ features to be corrupted. Let $M_{crpt}$ denote the subset of features to be corrupted. In the current literature, the $M_{crpt}$ features are \emph{randomly selected} from the $M$ features for each anchor to be corrupted. Furthermore, for each feature within the $M_{crpt}$ features, the replacement value is also \emph{randomly selected} from the feature column from the entire table. The augmentation function $g(\cdot)$ along with the contrastive learning procedure is described in \cref{alg:orig}. Note that mini-batches can be used within each epoch in the algorithm.
\begin{algorithm}
\caption{Contrastive Learning with Existing Corruption Procedure}
\label{alg:orig}
\small
\begin{algorithmic}[1]
\STATE Randomly initialize $\theta_e$ and $\theta_p$
\STATE Set $p$: percentage of features corrupted
\STATE Set $N_{eps}$: number of training epochs
\FOR {$e \gets  1$ to $N_{eps}$}
    \FOR{$i \gets 1$ to  $N_l+N_u$}
        \STATE Uniformly select $\lceil Mp \rceil$ features to form the set $\mathcal{M}_{crpt}$
        \STATE Initialize $\tilde{x}_i=x_i$.
        \FOR{$k \gets  1$ to $M $}
            \IF{$k \in \mathcal{M}_{crpt}$}
                \STATE Sample $j$ uniformly from $[N_l+N_u]$
                \STATE $\tilde{x}_i^{(k)} \xleftarrow{} x_j^{(k)}$
            \ENDIF
        \ENDFOR
        \STATE Compute and collect $z_i$ based on $x_i$ via \cref{equ:emb_orig}
        \STATE Compute and collect $\tilde{z}_i$ based on $\tilde{x}_i$ via \cref{equ:emb_aug}
    \ENDFOR
    \STATE Compute $\mathcal{L}$ via \cref{equ:contrastive_loss}
    \STATE Optimize $\theta_e$ and $\theta_p$ through gradient descent on $\mathcal{L}$
\ENDFOR
\end{algorithmic}
\end{algorithm}

%% file: method.tex
\section{Method}
\label{sec:method}
We formally propose two improvements over the existing tabular data augmentation procedure as described in \cref{sec:prob_III}. Specifically, we investigate in the sampling choices for both the replacement value on each feature, as well as the subset of features to be corrupted.

\subsection{How to Corrupt: Class-Conditioned Corruption}\label{sec:method_I}
Firstly, we focus on \emph{how to corrupt}, \textit{i.e.}, the selection of replacement values that are used to corrupt each selected feature. In contrastive learning, the views generated from each anchor are supposed to be semantically close to the anchor. Correspondingly, the augmentation function $g(\cdot)$ should preserve the semantics of the anchor data while generating variations. Nonetheless, the existing augmentation function described in \cref{alg:orig} is not necessarily semantic preserving, especially when the feature being corrupted is largely affecting the class identity of the row.

To improve the augmentation technique regarding this perspective, we propose to adopt the \emph{class-conditioned corruption} technique. In particular, when corrupting each selected feature in the anchor, we only sample the feature value from rows that are under the same class as the anchor. Given that the row used for corruption and the anchor belong to the same class, the semantics of their features are more likely to be similar, at least under the perspective of the down-stream classification task. 

The main challenge within the class-conditioned corruption process is to obtain class labels over the entire table. Conventionally, as elaborated in \cref{sec:prob_II}, contrastive learning is a self-supervised learning approach that does not require any label. However, contrastive learning is only responsible for training the encoder module of the neural network that leads to a semantically meaningful embedding space, which is rarely utilized in a standalone manner. In most applications, there exists a down-stream application for which the model will be fine-tuned on, at which point the target information is required. Therefore, the assumption of semi-supervised learning, where a small set of labels is available, is more realistic and more frequently encountered in real-world applications. Given the small set of class labels available under semi-supervised learning, it is a wasted opportunity to not incorporate such class information into the contrastive learning procedure, which is the case in existing tabular augmentation approaches.

Under the semi-supervised learning setting, we adapt the popular \emph{pseudo labeling} approach \citet{pseudolabel} for obtaining labels on the unlabeled training set. Specifically, a model is first trained on the small labeled set, and then utilized to run inference on the remaining unlabeled set. The model outputs are regarded as the pseudo targets for the unlabeled data. In this paper, we integrate this pseudo labeling process into the contrastive learning pre-training stage, through \emph{iteratively} updating the encoder to minimize the contrastive loss; and updating the classification head to minimize the supervised loss on the small labeled set for pseudo labels. \cref{alg:improved_how} outlines this process.

\begin{algorithm}[h]
\caption{Improved Corruption Procedure with Class-Conditioned Corruption}
\label{alg:improved_how}
\small
\begin{algorithmic}[1]
\STATE Randomly initialize $\theta_e$, $\theta_p$, and $\theta_c$
\STATE Set $p$: percentage of features corrupted
\STATE Set $N_{eps}$: number of training epochs
\STATE Set $N_{up}$: interval length for updating $\theta_c$
\FOR{$e \gets  1$ to $N_{eps}$}
    \FOR{$i \gets N_l+1$ to  $N_l+N_u$}
        \STATE Obtain pseudo labels: $c_i=f_{\theta_c}(f_{\theta_e}(x_i))$
    \ENDFOR
    \FOR{$i \gets 1$ to $N_l+N_u$}
        \STATE Uniformly select $\lceil Mp \rceil$ features to form the set $\mathcal{M}_{crpt}$
        \STATE Initialize $\tilde{x}_i=x_i$.
        \FOR{$k \gets  1$ to $M $}
            \IF{$k \in \mathcal{M}_{crpt}$}
                \STATE Let $\mathcal{N}_{c_i}\in[N_l+N_u]$ denote indices of samples under class $c_i$
                \STATE Sample $j$ uniformly from $\mathcal{N}_{c_i}$.
                \STATE $\tilde{x}_i^{(k)} \xleftarrow{} x_j^{(k)}$
            \ENDIF
        \ENDFOR
        \STATE Compute and collect $z_i$ based on $x_i$ via \cref{equ:emb_orig}.
        \STATE Compute and collect $\tilde{z}_i$ based on $\tilde{x}_i$ via \cref{equ:emb_aug}.
    \ENDFOR
    \STATE Compute $\mathcal{L}$ via \cref{equ:contrastive_loss}
    \STATE Optimize $\theta_e$ and $\theta_p$ through gradient descent on $\mathcal{L}$
    \IF{$e\mod{N_{up}} == 0$}
        \STATE Freeze $\theta_e$
        \STATE Optimize $\theta_c$ on the labeled data (via conventional supervised training)
    \ENDIF
\ENDFOR
\end{algorithmic}
\end{algorithm}

We note that there have been a couple of other approaches in the literature that explore incorporating class information into the augmentation procedure \citet{contramix,transtab}. The main difference from these two compared to our method proposed in this sub-section is the contrastive learning objective: both of these approaches utilize a \emph{supervised contrastive learning} objective \citet{supervisedcontra}, which is different from the classical contrastive learning objective as in \cref{equ:contrastive_loss}. In supervised contrastive learning, views belonging to the same class are pushed altogether, which would lead to less granular latent representations especially when the number of classes is low, which is often the case for tabular datasets.

\subsection{Where to Corrupt: Correlation-Based Feature Masking}\label{sec:method_II}

As the second direction of improvement, we explore the problem of \emph{where to corrupt}, \textit{i.e.}, which subset of features to corrupt. Instead of randomly selecting the subset of features to corrupt on each anchor, we hypothesize that incorporating the correlation information among features can improve the performance of contrastive learning. 

The intuition behind correlation-based feature masking can be better understood by drawing the analogy to image corruption. If we consider pixels of an image as its features, in general, adjacent pixels are correlated features, while the disjoint pixels that are far away apart are uncorrelated features. In the literature, image corruption has been explored as a data augmentation technique, where patches of an anchor image are corrupted. This technique can be regarded as selecting a subset of features that are \emph{highly correlated} for corruption. Conversely, selecting features that are \emph{highly uncorrelated} corresponds to corrupting sparsely located pixels on an image. The rationale behind this strategy follows that with each feature being corrupted, there exists correlated features left intact. Through reconstructing the corrupted features based on their correlated features, the model is encouraged to learn and utilize the knowledge of feature correlations, which will be helpful in performing downstream tasks.

The choice of the feature correlation metric is not trivial. The most straight-forward selection is by computing the covariance matrix. This metric however has two shortcomings: being incompatible to categorical features; and being limited to modeling only linear relationships. In this paper, we adopt a more general and flexible correlation measure, which is through fitting an XGBoost model \citet{xgboost} and obtaining the feature importance scores as a proxy for feature correlation. Specifically, given the entire training set of tabular data, for each feature, we fit an XGBoost model to predict this feature (classification for categorical features; regression for numerical features) based on the remaining features. We then utilize the normalized feature importance scores as the indicator on how each of the remaining feature correlates to the feature to be predicted. We relegate the description for the sampling procedure based on these feature importance scores to \cref{sec:app_I}. The detailed description of the entire correlation-based feature masking approach is shown in \cref{alg:improved_where}. Note that by directly using the XGBoost model feature importance scores, \cref{alg:improved_where} will provide us with subsets including highly-correlated features. For sampling sets of least-correlated features, we may simply use the negative values of the feature importance scores in \cref{alg:improved_where}.

\begin{algorithm}
\caption{Improved Corruption Procedure with Correlation-Based Feature Masking}
\label{alg:improved_where}
\small
\begin{algorithmic}[1]
\STATE Randomly initialize $\theta_e$ and $\theta_p$
\STATE Set $p$: percentage of features corrupted
\STATE Set $N_{eps}$: number of training epochs
\FOR{$k \gets 1$ to $M$}
    \STATE Train an XGBoost model to predict feature $k$ based on the remaining $M{-}1$ features over the training set $[N_l+N_u]$. 
    \STATE Collect $\mathbf{q}_k\in\mathcal{R}^{M{-}1}$: feature importance scores to $k$-th feature.
\ENDFOR
\FOR{$e \gets 1$ to $N_{eps}$}
    \FOR{$i \gets 1$ to  $N_l+N_u$}
        \STATE Sample $\lceil Mp \rceil$ features to form the set $\mathcal{M}_{crpt}$, as described in \cref{alg:sample_features} based on $\{\mathbf{q}_k, \forall k\}$.
        \STATE Initialize $\tilde{x}_i=x_i$.
        \FOR{$k \gets  1$ to $M $}
            \IF{$k \in \mathcal{M}_{crpt}$}
                \STATE Sample $j$ uniformly from $[N_l+N_u]$.
                \STATE $\tilde{x}_i^{(k)} \xleftarrow{} x_j^{(k)}$
            \ENDIF
        \ENDFOR
        \STATE Compute and collect $z_i$ based on $x_i$ via \cref{equ:emb_orig}
        \STATE Compute and collect $\tilde{z}_i$ based on $\tilde{x}_i$ via \cref{equ:emb_aug}
    \ENDFOR
    \STATE Compute $\mathcal{L}$ via \cref{equ:contrastive_loss}
    \STATE Optimize $\theta_e$ and $\theta_p$ through gradient descent on $\mathcal{L}$
\ENDFOR
\end{algorithmic}
\end{algorithm}

We note that the proposed class-conditioned corruption and correlation-based feature masking are two parallel improvements that can be implemented simultaneously, which corresponds to merging \cref{alg:improved_how} and \cref{alg:improved_where} together.

%% file: experiments.tex
\section{Experiments}
\label{sec:exp}
\subsection{Experimental Setup}\label{sec:exp_I}
We focus on the semi-supervised tabular classification task as described in \cref{sec:prob_I}. We adopt the contrastive learning approach for obtaining the pre-trained model encoder $\theta_e$, followed by training the classification head $\theta_c$. Within the contrastive learning process, we implement our proposed augmentation techniques as elaborated in \cref{alg:improved_how} and \cref{alg:improved_where}. To examine the effectiveness of both of our approaches, we compare them against the following methods:
\begin{itemize}
    \item No Pre-train: after randomly initializing the encoder, we freeze it and directly train the classification head on the labeled dataset. 
    \item Conventional Corruption: we perform contrastive learning using the conventional data augmentation approach as introduced in \cref{sec:prob_III}.
    \item Oracle Corruption: assume accurate targets are available over the entire training set.
\end{itemize}
We note that the \emph{conventional corruption} baseline essentially resembles the approach in \citet{scarf}. Nonetheless, as we could not replicate results in a stable manner following the settings specified in their paper (mostly due to their early-stopping setup), the hyper-parameters adopted in our simulations are slightly different from those in \citet{scarf}. Furthermore, we also note that the \emph{oracle corruption} baseline is not realistic under the semi-supervised learning assumption. This baseline serves to show an upper-bound on the performance improvement obtained by incorporating class-conditioned corruption into the data augmentation process. 

In terms of the neural network architecture specification, we implement all three modules, namely the encoder, the pre-train head, and the decoder, using fully-connected layers, with each of the hidden layers size of 256. We use 4 hidden layers for the encoder (\textit{i.e.}, $\theta_e$),  1 hidden layer and 1 output layer for the pre-train head (i.e., $\theta_p$), and 1 hidden layer and 1 output layer for the classification head (\textit{i.e.}, $\theta_c$). In \cref{tab:n_epochs}, we list out the number of epochs for each training stage.

\begin{table}[t]
    \addtocounter{footnote}{1}
    \centering
    \begin{tabular}{c|c|c}
    \hline
    Stage & Model Parameters & Epochs  \\
    \hline
    \makecell{Contrastive-learning \\ Pretraining} & $\theta_e$ \& $\theta_p$ & 500 \\
    \hline
    \makecell{Down-stream \\ Fine-Tuning} & $\theta_c$ & 100 \\
    \hline
    \makecell{Pseudo-labeling \\ Iterative Update\footnotemark} & $\theta_c$ & 10 \\
    \hline
    \end{tabular}
    \caption{Training Set-ups at Different Stages.}
    \label{tab:n_epochs}
\end{table}
\footnotetext[2]{$N_{up}$ in \cref{alg:improved_how}.}

\subsection{Contrastive Loss Learning Curve \& Embedding Visualization}
\begin{figure}[t]
     \centering
     \includegraphics[width=\linewidth]{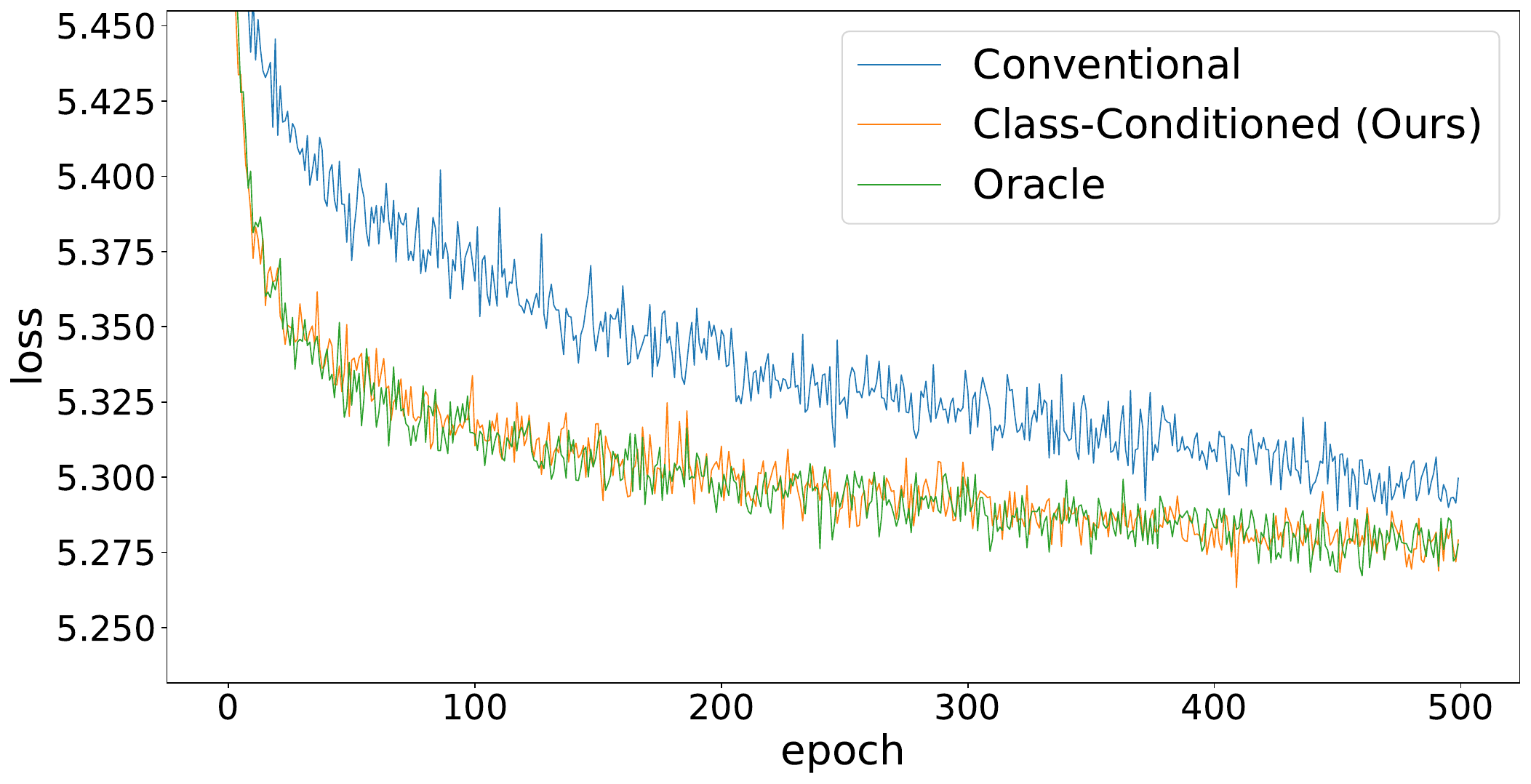}
     \vspace{-1.5em}
     \caption{Contrastive Loss Learning Curves for Competing Augmentation Methods in the Pre-training Stage. Compared to the conventional tabular augmentation (corruption with randomly sampled values), our method achieves noticeably lower contrastive loss at a faster pace. Note that our method matches the loss optimization results by the oracle corruption method (which has the knowledge of all labels).}
     \label{fig:learn_curve}
\end{figure}

\begin{figure*}[t!]
     \centering
     \includegraphics[width=\linewidth]{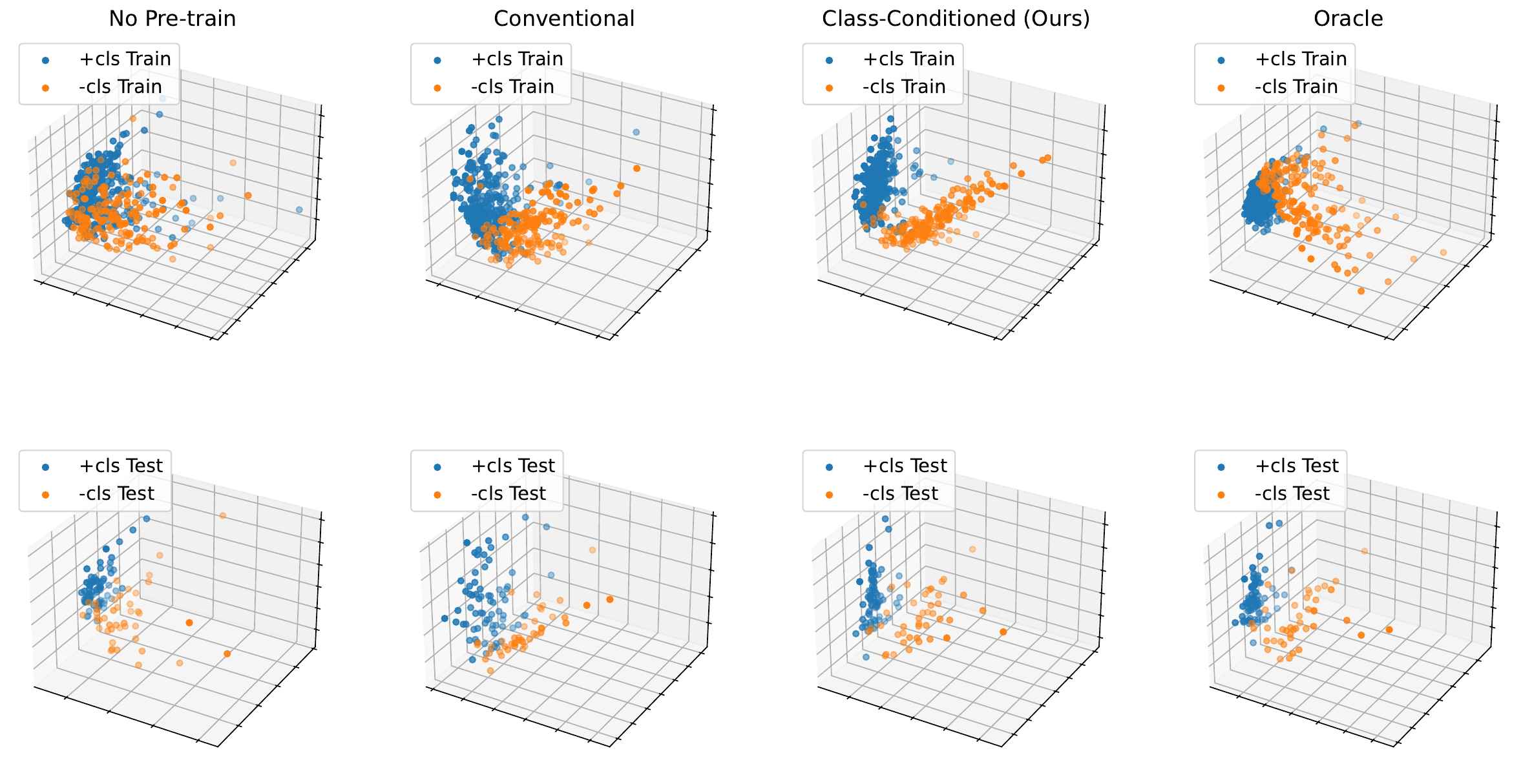}
     \vspace{-1.5em}
     \caption{Learned Embeddings from Augmentation Methods. The embeddings are computed by the encoder pre-trained under each method, and are visualized in 3-D space through dimensionality reduction with PCA. Evidently, after pre-training under our approach, the separation between samples (both train and test samples) from different classes is more prominent in the learned embedding space compared to no pre-training or pre-training under the conventional augmentation approach.}
     \label{fig:embed}
\end{figure*}

We first examine the contrastive learning progress as well as the resultant embedding space from the model under each augmentation technique as introduced in \cref{sec:exp_I}. For illustration simplicity, in this sub-section, we use the simple tabular classification dataset ``breast cancer" provided in the sklearn package. We first plot the contrastive loss value as defined in \cref{equ:contrastive_loss} in \cref{fig:learn_curve}. As illustrated, introducing the class information into the tabular augmentation process indeed encourages the views created to be more similar to the anchors, leading to a noticeably lower contrastive loss.

We then visualize the learned embedding space under each method. After the contrastive learning pre-training, we take the embeddings computed by the encoder, \textit{i.e.}, $f_{\theta_e}(x)$, and visualize how these embeddings are distributed across different classes. We apply principal component analysis (PCA) to reduce the learned embeddings down to 3 dimensions, and plot them by class identities in \cref{fig:embed}. As evident from the visualizations, compared to the conventional augmentation technique, the proposed class-conditioned corruption approach helps to learn an embedding space where there is more separation between data from different classes, for both the training set and the testing set. Therefore, it is easier to learn and perform classification based on such an embedding space.

\subsection{Classification Performances with Class-Conditioned Corruption}

We evaluate the down-stream classification performance of each method over a large set of tabular data from the OpenML-CC18 dataset \citet{openmlcc}. We follow the specifications for training and evaluating the models as listed in \cref{sec:exp_I}. For categorical features, we use one-hot encoding \emph{after} the corruption step to ensure the representation of the tabular data always follows the valid encoding format. 

\begin{figure}[h!]
     \centering
     \includegraphics[width=6.5cm]{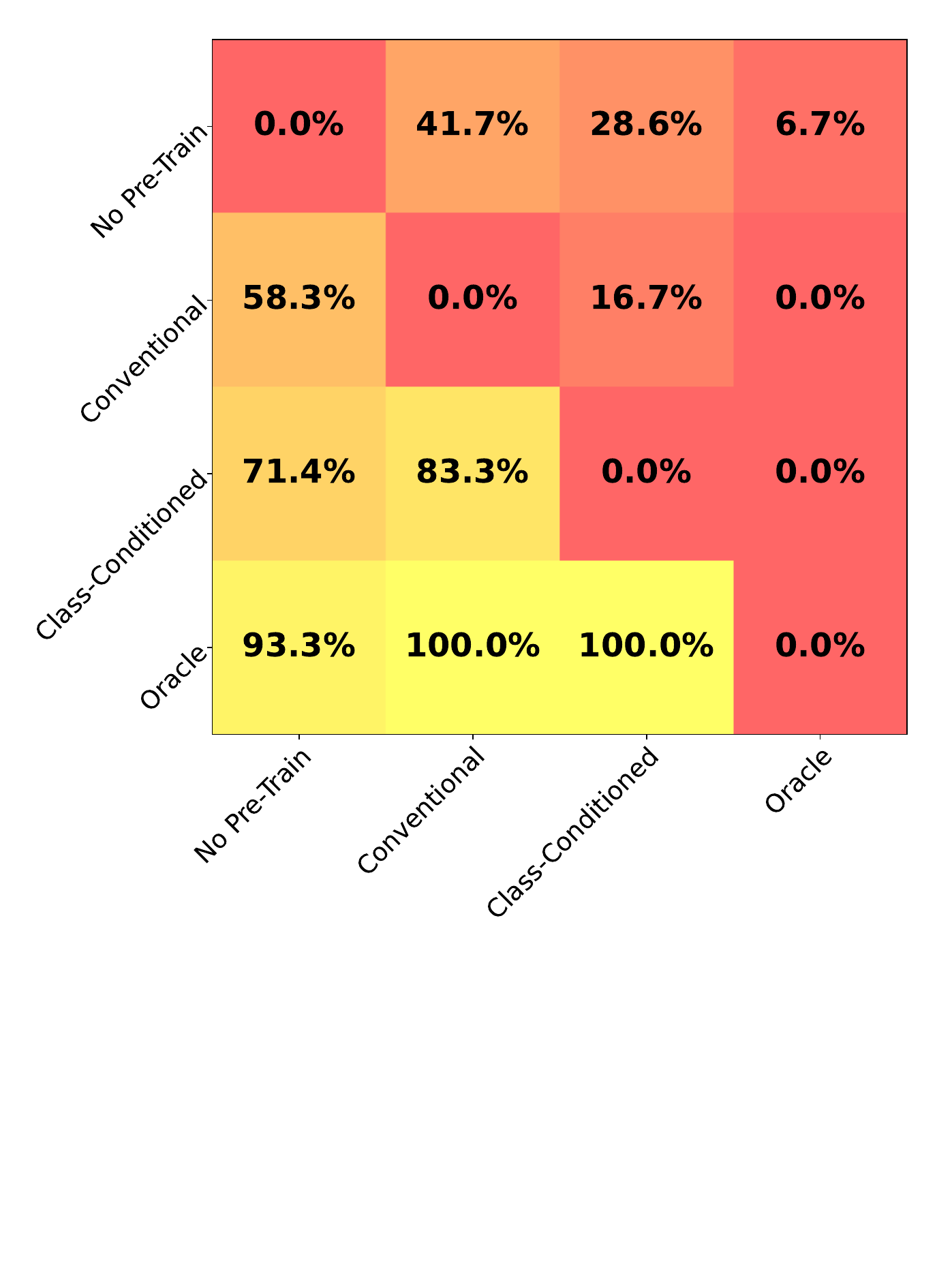}
     \vspace{-7em}
     \caption{Classification Accuracy Win Matrix among competing augmentation methods on \textit{How to Corrupt}. Our method achieves better classification results compared to the conventional approach over a large portion of datasets.}
     \label{fig:win_mat_accuracy}
\end{figure}

We present the win matrix $W$ on classification accuracies among each competing method in \cref{fig:win_mat_accuracy}, with the ($i$, $j$) entry of the matrix being the ratio computed over the 30 OpenML-CC18 tables as follows:
\begin{align}
    W_{i,j}=\frac{\sum\limits_{d=1}^{30}\mathds{1}[\text{method $i$ beats $j$ on dataset $d$}]}{\sum\limits_{d=1}^{30}\mathds{1}[\text{method $i$ beats OR loses to $j$ on dataset $d$}]}
\end{align}
We use a t-test with unequal population variances with a statistical significance level (i.e. the $p$-value) of 0.05, and only compute the ratios over the datasets from which statistical significant comparison results can be obtained. We relegate the full classification accuracy results, as well as full AUROC (area under receiver operating characteristic curve) results, in \cref{tab:accuracy-results-how} and \cref{tab:auroc-results-how} respectively in \cref{sec:app_II}. 

As evident through the comparison, our proposed class-conditioned corruption approach shows noticeable improvements over the conventional corruption method, where our approach achieves improvements on 83\% of the datasets. Furthermore, the oracle method shows dominant performance over the rest of the methods on both the accuracy and AUROC metrics, further validating the benefits of considering class identity information into the tabular corruption process.

\subsection{Limitations from Corruption-Based Augmentation}
One interesting observation we would like to note is that, to much of our surprise, the classification results from no pre-train are occasionally higher than the results achieved after pre-training (even including the oracle). This observation potentially reveals that for certain tabular datasets, conducting contrastive learning via any corruption-based augmentation has its intrinsic limitations in general. We believe this finding could motivate further investigations into whether or not performing corruption for data augmentation is always beneficial in the tabular domain. 

One comprehensible example for our argument is the \emph{Balance Scale} table in the OpenML-CC18 datasets. In the Balance Scale table, each row consists of weights and distances to a pivot in a two-ended scale. Naturally, when we corrupt by swapping weights or distances, the scale balance can change abruptly, leading to a generated row that contradicts the laws of physics. We do however emphasize that such cases are likely rare in more complex and realistic datasets.

\subsection{Classification Performances with Correlation-Based Feature Masking}

For correlation-based feature masking, we explore both variants as elaborated in \cref{sec:method_II}: sampling features that are highly correlated, and sampling features that are highly uncorrelated. We experiment each of these variants in conjunction with our class-conditioned corruption method. Specifically, for each feature masking approach with our class-conditioned corruption technique, we follow the procedure by merging \cref{alg:improved_how} and \cref{alg:improved_where} together.

The classification accuracy results as well as the AUROC results of the corresponding methods on the down-stream classification task over the OpenML-CC18 datasets are shown in \cref{tab:accuracy-results-where} and \cref{tab:auroc-results-where} in \cref{sec:app_III}, respectively. As shown by the results, there lacks consistent improvement from exploiting feature correlations when corrupting tabular data. We believe that this is largely due to the fact that from careful processing and preparation of the tabular datasets in OpenML-CC18, much of the redundancy has been removed in each table, resulting in the set of features being largely independent and sharing little correlation within each other. We refer to the Balance Scale table again as a clear example, in which each feature is completely independent from all other features.  

In an attempt to understand the strength of feature correlations within each table, we have computed and presented the \textit{feature correlation value range} for each table in \cref{tab:accuracy-results-where} and \cref{tab:auroc-results-where}. The feature correlation value range is the range of feature correlation values computed for each table as described in \cref{sec:method_II}. However, we do emphasize that the correlation value range only provides the tip of the iceberg in terms of describing the full feature correlation structure within a given table. A potential direction of future research is to better quantify the feature correlation structure and strength within tabular data, which we believe will also benefit the data augmentation process such as this selective feature masking approach. 

Essentially, there is insignificant feature correlation structure in these benchmark tables to start with that can be exploited in the data augmentation process. We hypothesize that our correlation-based feature masking approach has the potential for improving the contrastive learning over larger, more complex, and more crude datasets that are frequently encountered in real-world applications.

%% file: appendices.tex
\appendix
\onecolumn
\section{Sampling Procedure for Correlation-Based Feature Masking}\label{sec:app_I}

In our proposed correlation-based feature masking approach as in \cref{sec:method_II}, the subset of features to be corrupted on each anchor row is sampled conditioned on the feature correlations. In this appendix section, we provide in details for this sampling procedure. 

Firstly, as described in \cref{alg:improved_where}, for a table with $M$ features, we train an XGBoost model for predicting each feature based on the remaining $M{-}1$ features. We then collect $M$ sets of feature importance values, which we denote by $\{\mathbf{q}_k\}$ for $k\in\{1,\dots,M\}$, such that $\mathbf{q}_k^{(j)}$ denotes the feature importance of the $j$-th feature in predicting the $k$-th feature.

Based on the set of feature importance values, we sample the subset of features for each anchor independently, through the procedure as detailed in \cref{alg:sample_features}.

\begin{algorithm}
\caption{Sampling Procedure for the Feature Subset Based on Feature Correlation Values}
\label{alg:sample_features}
\small
\begin{algorithmic}[1]
\INPUT $\{\mathbf{q}_k\}$ for $k\in\{1,\dots,M\}$
\INPUT $M$: the number of features
\INPUT $p$: the percentage of features to be corrupted
\STATE Initialize $C_{corr}\in\mathcal{R}^{M\times M}$ as follows:
\vspace{0.5em}

$\begin{aligned}
    &C_{corr}[i,i] = 0 \quad \forall i \in [M] \\
    &C_{corr}[i,j] = \mathbf{q}_i^{(j)} \quad \forall i\in[M],j\in[M],i\neq j		\hskip 0.36\textwidth \text{if sampling \emph{most correlated} features} \\
    &C_{corr}[i,j] = -\mathbf{q}_i^{(j)} \quad \forall i\in[M],j\in[M],i\neq j		\hskip 0.345\textwidth \text{if sampling \emph{least correlated} features}
\end{aligned}$
\vspace{0.35em}
\STATE Uniformly sample $f_1$ from the index set $[M]$.
\STATE Initialize the set $\mathcal{M}_{crpt} \gets [f_1]$. 
\vspace{0.2em}
\FOR{$i \gets 2$ to $\lceil Mp \rceil$}
    \vspace{0.2em}
    \STATE Select the sub-matrix $C_{corr}^{\text{part}}\gets C_{corr}[\mathcal{M}_{crpt},:]$ \hfill (Select the set of rows from $C_{corr}$ indexed by $\mathcal{M}_{crpt}$)
    \vspace{0.2em}
    \STATE $\mathbf{p}_i\gets min(C_{corr}^{\text{part}}, \text{axis}{=}1)$ \hfill (Select the column-wise minimum values)
    \STATE $\mathbf{p}_i^{(k)}\gets0\quad\forall k\in\mathcal{M}_{crpt}$ \hfill (Set probabilities for already selected features to zero)
    \vspace{0.2em}
    \STATE $\mathbf{\hat{p}}_i\gets \frac{\mathbf{p}_i}{\sum_k\left(\mathbf{p}_i^{(k)}\right)}$
    \STATE Sample the feature index $f_i\sim\mathbf{\hat{p}}_i$.
    \STATE Append $f_i$ to $\mathcal{M}_{crpt}$.
\ENDFOR
\STATE \textbf{Return} $\mathcal{M}_{crpt}$
\end{algorithmic}
\end{algorithm}

Note that in obtaining the unnormalized sampling probability vector $\mathbf{p}_i$ in \cref{alg:sample_features}, we take the minimum among feature importance values among the columns (i.e. along the features) of the sub-matrix $C_{corr}^{\text{part}}$. This minimization step allows us to collect a new feature that is most correlated (or least correlated) to each of the already selected features in the set $\mathcal{M}_{crpt}$. 

\newpage
\section{Classification Accuracies on How to Corrupt under Random Feature Selection}\label{sec:app_II}

\begin{table*}[h]
\small
\caption{Semi-Supervised Classification Accuracies (in Percentages) on \textit{How to Corrupt}. For each method (except for No-PreTrain), the features corrupted are randomly selected.} 
\centering
\begin{tabular}{c|ccccc}
\hline 
 Datasets (DID) & No-PreTrain & Random & Class (ours) & Oracle \\ 
\hline 
balance-scale (11) & $90.00\pm 0.66$ & $90.50\pm 0.93$ & $\mathbf{91.60\pm 0.97}$ & $94.50\pm 0.67$ \\ 
mfeat-fourier (14) & $68.00\pm 0.88$ & $78.19\pm 0.96$ & $\mathbf{79.22\pm 0.79}$ & $81.88\pm 0.58$ \\ 
breast-w (15) & $95.98\pm 0.56$ & $96.79\pm 0.40$ & $96.70\pm 0.36$ & $96.61\pm 0.30$ \\ 
mfeat-karhunen (16) & $81.75\pm 0.68$ & $95.22\pm 0.43$ & $94.50\pm 0.28$ & $97.19\pm 0.26$ \\ 
mfeat-morphological (18) & $72.62\pm 0.48$ & $70.62\pm 0.58$ & $\mathbf{71.19\pm 0.33}$ & $73.59\pm 0.73$ \\ 
mfeat-zernike (22) & $76.53\pm 0.44$ & $76.22\pm 0.65$ & $\mathbf{78.78\pm 0.46}$ & $82.44\pm 0.51$ \\ 
cmc (23) & $53.43\pm 0.95$ & $50.51\pm 0.94$ & $\mathbf{51.53\pm 1.40}$ & $50.21\pm 1.03$ \\ 
credit-approval (29) & $82.25\pm 1.03$ & $80.89\pm 1.30$ & $\mathbf{83.42\pm 1.26}$ & $84.51\pm 1.23$ \\ 
credit-g (31) & $71.31\pm 0.40$ & $71.25\pm 1.14$ & $\mathbf{72.88\pm 1.35}$ & $75.56\pm 0.72$ \\ 
diabetes (37) & $74.03\pm 0.74$ & $72.89\pm 1.06$ & $\mathbf{74.76\pm 0.91}$ & $73.78\pm 1.09$ \\ 
tic-tac-toe (50) & $78.39\pm 0.70$ & $88.22\pm 0.80$ & $85.68\pm 1.31$ & $88.87\pm 0.60$ \\ 
vehicle (54) & $72.06\pm 0.91$ & $71.10\pm 1.11$ & $\mathbf{73.24\pm 0.80}$ & $74.34\pm 1.09$ \\ 
eucalyptus (188) & $54.56\pm 1.27$ & $54.31\pm 1.07$ & $\mathbf{58.36\pm 1.07}$ & $61.91\pm 0.97$ \\ 
analcatdata-authorship (458) & $94.67\pm 0.47$ & $99.85\pm 0.14$ & $99.70\pm 0.10$ & $99.63\pm 0.15$ \\ 
analcatdata-dmft (469) & $20.08\pm 1.06$ & $18.67\pm 0.93$ & $\mathbf{19.45\pm 0.70}$ & $19.61\pm 0.86$ \\ 
pc4 (1049) & $89.43\pm 0.40$ & $85.92\pm 0.76$ & $\mathbf{87.97\pm 0.54}$ & $89.00\pm 0.62$ \\ 
pc3 (1050) & $88.74\pm 0.29$ & $89.26\pm 0.49$ & $88.70\pm 0.34$ & $88.78\pm 0.46$ \\ 
kc2 (1063) & $83.57\pm 0.62$ & $82.86\pm 0.58$ & $\mathbf{83.45\pm 0.67}$ & $83.21\pm 0.77$ \\ 
pc1 (1068) & $92.51\pm 0.31$ & $92.91\pm 0.21$ & $91.84\pm 0.64$ & $93.13\pm 0.07$ \\ 
banknote-authentication (1462) & $99.64\pm 0.09$ & $99.64\pm 0.17$ & $99.05\pm 0.23$ & $99.41\pm 0.16$ \\ 
blood-transfusion-service-center (1464) & $78.75\pm 0.36$ & $76.92\pm 0.91$ & $\mathbf{77.67\pm 1.02}$ & $77.67\pm 0.93$ \\ 
ilpd (1480) & $68.16\pm 1.54$ & $68.06\pm 1.06$ & $\mathbf{68.27\pm 1.07}$ & $70.09\pm 1.36$ \\ 
qsar-biodeg (1494) & $83.41\pm 1.11$ & $84.06\pm 0.71$ & $83.59\pm 0.62$ & $86.97\pm 0.94$ \\ 
wdbc (1510) & $95.61\pm 0.68$ & $96.60\pm 0.61$ & $\mathbf{97.15\pm 0.37}$ & $97.92\pm 0.44$ \\ 
cylinder-bands (6332) & $63.89\pm 1.39$ & $70.02\pm 1.10$ & $68.63\pm 1.24$ & $74.07\pm 1.01$ \\ 
dresses-sales (23381) & $55.12\pm 0.91$ & $51.62\pm 1.36$ & $50.88\pm 1.79$ & $53.00\pm 1.41$ \\ 
MiceProtein (40966) & $74.25\pm 1.98$ & $79.11\pm 1.16$ & $\mathbf{89.58\pm 0.73}$ & $95.83\pm 0.50$ \\ 
car (40975) & $85.51\pm 0.77$ & $93.21\pm 0.25$ & $93.10\pm 0.53$ & $96.17\pm 0.44$ \\ 
steel-plates-fault (40982) & $70.73\pm 0.67$ & $68.32\pm 0.67$ & $\mathbf{70.98\pm 0.95}$ & $72.78\pm 0.55$ \\ 
climate-model-simulation-crashes (40994) & $92.48\pm 0.30$ & $92.59\pm 0.46$ & $91.55\pm 0.62$ & $93.40\pm 0.42$ \\ 
\hline 
\end{tabular}
\label{tab:accuracy-results-how}
\end{table*}
In \cref{tab:accuracy-results-how} above, the results where our proposed class-conditioned corruption approach outperforms the conventional random corruption approach with concrete margins are marked in bold fonts. In the majority of the datasets, the class-conditioned approach shows improvements over the conventional approach, while the two methods show close to equal performances for the most of the remaining datasets.

\clearpage
\begin{table*}[h]
\small
\caption{Semi-Supervised Classification AUROCs on \textit{How to Corrupt}. For each method (except for No-PreTrain), the features corrupted are randomly selected.} 
\centering
\begin{tabular}{c|ccccc}
\hline 
 Datasets (DID) & No-PreTrain & Random & Class & Oracle \\ 
\hline 
balance-scale (11) & $0.95\pm 0.00$ & $0.95\pm 0.01$ & $\mathbf{0.97\pm 0.01}$ & $0.99\pm 0.00$ \\ 
mfeat-fourier (14) & $0.94\pm 0.00$ & $0.97\pm 0.00$ & $0.97\pm 0.00$ & $0.98\pm 0.00$ \\ 
breast-w (15) & $0.99\pm 0.00$ & $0.99\pm 0.00$ & $0.99\pm 0.00$ & $0.99\pm 0.00$ \\ 
mfeat-karhunen (16) & $0.98\pm 0.00$ & $1.00\pm 0.00$ & $1.00\pm 0.00$ & $1.00\pm 0.00$ \\ 
mfeat-morphological (18) & $0.96\pm 0.00$ & $0.96\pm 0.00$ & $0.96\pm 0.00$ & $0.96\pm 0.00$ \\ 
mfeat-zernike (22) & $0.97\pm 0.00$ & $0.97\pm 0.00$ & $0.97\pm 0.00$ & $0.98\pm 0.00$ \\ 
cmc (23) & $0.70\pm 0.01$ & $0.68\pm 0.01$ & $\mathbf{0.69\pm 0.01}$ & $0.69\pm 0.01$ \\ 
credit-approval (29) & $0.89\pm 0.01$ & $0.88\pm 0.01$ & $\mathbf{0.90\pm 0.01}$ & $0.92\pm 0.01$ \\ 
credit-g (31) & $0.69\pm 0.01$ & $0.74\pm 0.01$ & $0.74\pm 0.01$ & $0.78\pm 0.01$ \\ 
diabetes (37) & $0.80\pm 0.01$ & $0.79\pm 0.01$ & $0.78\pm 0.02$ & $0.81\pm 0.01$ \\ 
tic-tac-toe (50) & $0.84\pm 0.01$ & $0.94\pm 0.00$ & $0.92\pm 0.01$ & $0.96\pm 0.00$ \\ 
vehicle (54) & $0.91\pm 0.00$ & $0.89\pm 0.01$ & $\mathbf{0.91\pm 0.00}$ & $0.92\pm 0.00$ \\ 
eucalyptus (188) & $0.84\pm 0.00$ & $0.79\pm 0.01$ & $\mathbf{0.83\pm 0.01}$ & $0.86\pm 0.01$ \\ 
analcatdata-authorship (458) & $1.00\pm 0.00$ & $1.00\pm 0.00$ & $1.00\pm 0.00$ & $1.00\pm 0.00$ \\ 
analcatdata-dmft (469) & $0.54\pm 0.01$ & $0.52\pm 0.01$ & $\mathbf{0.53\pm 0.01}$ & $0.53\pm 0.01$ \\ 
pc4 (1049) & $0.87\pm 0.01$ & $0.82\pm 0.01$ & $\mathbf{0.84\pm 0.01}$ & $0.88\pm 0.01$ \\ 
pc3 (1050) & $0.77\pm 0.02$ & $0.80\pm 0.02$ & $0.79\pm 0.02$ & $0.82\pm 0.02$ \\ 
kc2 (1063) & $0.80\pm 0.02$ & $0.80\pm 0.02$ & $0.78\pm 0.02$ & $0.79\pm 0.03$ \\ 
pc1 (1068) & $0.74\pm 0.03$ & $0.67\pm 0.04$ & $\mathbf{0.74\pm 0.02}$ & $0.74\pm 0.01$ \\ 
banknote-authentication (1462) & $1.00\pm 0.00$ & $1.00\pm 0.00$ & $1.00\pm 0.00$ & $1.00\pm 0.00$ \\ 
blood-transfusion-service-center (1464) & $0.74\pm 0.01$ & $0.69\pm 0.02$ & $0.69\pm 0.02$ & $0.69\pm 0.02$ \\ 
ilpd (1480) & $0.71\pm 0.02$ & $0.71\pm 0.01$ & $0.70\pm 0.02$ & $0.72\pm 0.02$ \\ 
qsar-biodeg (1494) & $0.89\pm 0.00$ & $0.89\pm 0.00$ & $0.89\pm 0.01$ & $0.92\pm 0.01$ \\ 
wdbc (1510) & $0.99\pm 0.00$ & $0.99\pm 0.00$ & $0.99\pm 0.00$ & $1.00\pm 0.00$ \\ 
cylinder-bands (6332) & $0.67\pm 0.02$ & $0.76\pm 0.01$ & $0.75\pm 0.01$ & $0.81\pm 0.01$ \\ 
dresses-sales (23381) & $0.56\pm 0.02$ & $0.49\pm 0.02$ & $0.49\pm 0.02$ & $0.52\pm 0.02$ \\ 
MiceProtein (40966) & $0.97\pm 0.00$ & $0.97\pm 0.00$ & $\mathbf{0.98\pm 0.00}$ & $1.00\pm 0.00$ \\ 
car (40975) & $0.96\pm 0.00$ & $0.99\pm 0.00$ & $0.99\pm 0.00$ & $1.00\pm 0.00$ \\ 
steel-plates-fault (40982) & $0.92\pm 0.00$ & $0.92\pm 0.00$ & $0.92\pm 0.01$ & $0.93\pm 0.00$ \\ 
climate-model-simulation-crashes (40994) & $0.85\pm 0.02$ & $0.86\pm 0.02$ & $0.85\pm 0.02$ & $0.91\pm 0.01$ \\ 
\hline  
\end{tabular}
\label{tab:auroc-results-how}
\end{table*}
In \cref{tab:auroc-results-how} above, the results where our proposed class-conditioned corruption approach outperforms the conventional random corruption approach with concrete margins are marked in bold fonts. 

\clearpage

\section{Classification Performances on Where to Corrupt for Correlation-Based Feature Masking}\label{sec:app_III}

\begin{table*}[h]
\small
\caption{Semi-Supervised Classification Accuracies (in Percentages) on \textit{Where to Corrupt}. For each method (except for No-PreTrain), the class-conditioned corruption strategy is used for sampling values used in the corruption.} 
\centering
\begin{tabular}{cc|ccc}
\hline 
 Datasets (DID) & \makecell{Feature Correlation \\ Value Range} & Random Features & Least Correlated & Most Correlated \\ 
\hline 
balance-scale (11) & 0.03 & $91.60\pm 0.97$ & $92.20\pm 0.94$ & $91.70\pm 0.84$ \\ 
mfeat-fourier (14) & 0.14 & $79.22\pm 0.79$ & $79.03\pm 0.69$ & $78.94\pm 0.66$ \\ 
breast-w (15) & 0.33 & $96.70\pm 0.36$ & $96.70\pm 0.44$ & $96.52\pm 0.41$ \\ 
mfeat-karhunen (16) & 0.07 & $94.50\pm 0.28$ & $94.84\pm 0.35$ & $94.97\pm 0.22$ \\ 
mfeat-morphological (18) & 0.72 & $71.19\pm 0.33$ & $70.50\pm 0.54$ & $71.16\pm 0.31$ \\ 
mfeat-zernike (22) & 0.51 & $78.78\pm 0.46$ & $78.56\pm 0.28$ & $78.81\pm 0.29$ \\ 
cmc (23) & 0.13 & $51.53\pm 1.40$ & $50.04\pm 0.72$ & $50.30\pm 0.76$ \\ 
credit-approval (29) & 0.29 & $83.42\pm 1.26$ & $83.51\pm 1.17$ & $83.42\pm 1.22$ \\ 
credit-g (31) & 0.11 & $72.88\pm 1.35$ & $72.88\pm 0.98$ & $74.25\pm 0.56$ \\ 
diabetes (37) & 0.22 & $74.76\pm 0.91$ & $73.46\pm 1.57$ & $73.78\pm 1.05$ \\ 
tic-tac-toe (50) & 0.04 & $85.68\pm 1.31$ & $86.72\pm 0.94$ & $87.30\pm 0.82$ \\ 
vehicle (54) & 0.44 & $73.24\pm 0.80$ & $71.32\pm 1.07$ & $72.21\pm 0.89$ \\ 
eucalyptus (188) & 0.38 & $58.36\pm 1.07$ & $51.69\pm 1.07$ & $52.53\pm 1.02$ \\ 
analcatdata-authorship (458) & 0.11 & $99.70\pm 0.10$ & $99.63\pm 0.10$ & $99.63\pm 0.21$ \\ 
analcatdata-dmft (469) & 0.18 & $19.45\pm 0.70$ & $18.91\pm 0.71$ & $19.38\pm 0.99$ \\ 
pc4 (1049) & 0.39 & $87.97\pm 0.54$ & $87.71\pm 0.59$ & $87.33\pm 0.69$ \\ 
pc3 (1050) & 0.37 & $88.70\pm 0.34$ & $89.38\pm 0.34$ & $89.42\pm 0.24$ \\ 
kc2 (1063) & 0.36 & $83.45\pm 0.67$ & $82.86\pm 0.77$ & $84.17\pm 0.95$ \\ 
pc1 (1068) & 0.36 & $91.84\pm 0.64$ & $92.91\pm 0.26$ & $93.13\pm 0.07$ \\ 
banknote-authentication (1462) & 0.25 & $99.05\pm 0.23$ & $99.23\pm 0.19$ & $99.09\pm 0.19$ \\ 
blood-transfusion-service-center (1464) & 0.48 & $77.67\pm 1.02$ & $76.25\pm 0.90$ & $75.92\pm 0.65$ \\ 
ilpd (1480) & 0.35 & $68.27\pm 1.07$ & $68.27\pm 0.67$ & $68.70\pm 0.87$ \\ 
qsar-biodeg (1494) & 0.34 & $83.59\pm 0.62$ & $83.89\pm 0.59$ & $84.06\pm 0.60$ \\ 
wdbc (1510) & 0.43 & $97.15\pm 0.37$ & $97.26\pm 0.42$ & $96.93\pm 0.35$ \\ 
cylinder-bands (6332) & 0.19 & $68.63\pm 1.24$ & $68.63\pm 1.17$ & $69.33\pm 1.17$ \\ 
dresses-sales (23381) & 0.05 & $50.88\pm 1.79$ & $51.00\pm 0.40$ & $51.50\pm 0.79$ \\ 
MiceProtein (40966) & 0.26 & $89.58\pm 0.73$ & $88.95\pm 1.15$ & $89.47\pm 0.90$ \\ 
car (40975) & 0.01 & $93.10\pm 0.53$ & $93.28\pm 0.70$ & $93.28\pm 0.43$ \\ 
steel-plates-fault (40982) & 0.55 & $70.98\pm 0.95$ & $71.18\pm 0.93$ & $71.14\pm 0.76$ \\ 
climate-model-simulation-crashes (40994) & 0.07 & $91.55\pm 0.62$ & $91.44\pm 0.48$ & $90.97\pm 0.46$ \\ 
\hline 
\end{tabular}
\label{tab:accuracy-results-where}
\end{table*} 

\clearpage

\begin{table*}[h]
\small
\caption{Semi-Supervised Classification AUROCs on \textit{Where to Corrupt}. For each method (except for No-PreTrain), the class-conditioned corruption strategy is used for sampling values used in the corruption.} 
\centering
\begin{tabular}{cc|cccc}
\hline 
 Datasets (DID) & \makecell{Feature Correlation \\ Value Range} & Random Features & Least Correlated & Most Correlated \\ 
\hline 
balance-scale (11) & 0.03 & $0.97\pm 0.01$ & $0.97\pm 0.00$ & $0.97\pm 0.01$ \\ 
mfeat-fourier (14) & 0.14 & $0.97\pm 0.00$ & $0.97\pm 0.00$ & $0.97\pm 0.00$ \\ 
breast-w (15) & 0.33 & $0.99\pm 0.00$ & $0.99\pm 0.00$ & $0.99\pm 0.00$ \\ 
mfeat-karhunen (16) & 0.07 & $1.00\pm 0.00$ & $1.00\pm 0.00$ & $1.00\pm 0.00$ \\ 
mfeat-morphological (18) & 0.72 & $0.96\pm 0.00$ & $0.96\pm 0.00$ & $0.96\pm 0.00$ \\ 
mfeat-zernike (22) & 0.51 & $0.97\pm 0.00$ & $0.97\pm 0.00$ & $0.97\pm 0.00$ \\ 
cmc (23) & 0.13 & $0.69\pm 0.01$ & $0.68\pm 0.00$ & $0.68\pm 0.00$ \\ 
credit-approval (29) & 0.29 & $0.90\pm 0.01$ & $0.90\pm 0.01$ & $0.90\pm 0.01$ \\ 
credit-g (31) & 0.11 & $0.74\pm 0.01$ & $0.74\pm 0.01$ & $0.75\pm 0.01$ \\ 
diabetes (37) & 0.22 & $0.78\pm 0.02$ & $0.78\pm 0.02$ & $0.78\pm 0.01$ \\ 
tic-tac-toe (50) & 0.04 & $0.92\pm 0.01$ & $0.93\pm 0.01$ & $0.93\pm 0.00$ \\ 
vehicle (54) & 0.44 & $0.91\pm 0.00$ & $0.90\pm 0.00$ & $0.90\pm 0.00$ \\ 
eucalyptus (188) & 0.38 & $0.83\pm 0.01$ & $0.77\pm 0.01$ & $0.78\pm 0.01$ \\ 
analcatdata-authorship (458) & 0.11 & $1.00\pm 0.00$ & $1.00\pm 0.00$ & $1.00\pm 0.00$ \\ 
analcatdata-dmft (469) & 0.18 & $0.53\pm 0.01$ & $0.52\pm 0.01$ & $0.53\pm 0.01$ \\ 
pc4 (1049) & 0.39 & $0.84\pm 0.01$ & $0.84\pm 0.01$ & $0.84\pm 0.01$ \\ 
pc3 (1050) & 0.37 & $0.79\pm 0.02$ & $0.80\pm 0.02$ & $0.80\pm 0.02$ \\ 
kc2 (1063) & 0.36 & $0.78\pm 0.02$ & $0.79\pm 0.02$ & $0.79\pm 0.02$ \\ 
pc1 (1068) & 0.36 & $0.74\pm 0.02$ & $0.62\pm 0.05$ & $0.67\pm 0.02$ \\ 
banknote-authentication (1462) & 0.25 & $1.00\pm 0.00$ & $1.00\pm 0.00$ & $1.00\pm 0.00$ \\ 
blood-transfusion-service-center (1464) & 0.48 & $0.69\pm 0.02$ & $0.68\pm 0.02$ & $0.68\pm 0.02$ \\ 
ilpd (1480) & 0.35 & $0.70\pm 0.02$ & $0.70\pm 0.01$ & $0.69\pm 0.01$ \\ 
qsar-biodeg (1494) & 0.34 & $0.89\pm 0.01$ & $0.89\pm 0.01$ & $0.89\pm 0.01$ \\ 
wdbc (1510) & 0.43 & $0.99\pm 0.00$ & $1.00\pm 0.00$ & $1.00\pm 0.00$ \\ 
cylinder-bands (6332) & 0.19 & $0.75\pm 0.01$ & $0.76\pm 0.01$ & $0.76\pm 0.01$ \\ 
dresses-sales (23381) & 0.05 & $0.49\pm 0.02$ & $0.50\pm 0.01$ & $0.50\pm 0.02$ \\ 
MiceProtein (40966) & 0.26 & $0.98\pm 0.00$ & $0.98\pm 0.00$ & $0.98\pm 0.00$ \\ 
car (40975) & 0.01 & $0.99\pm 0.00$ & $0.99\pm 0.00$ & $0.99\pm 0.00$ \\ 
steel-plates-fault (40982) & 0.55 & $0.92\pm 0.01$ & $0.92\pm 0.00$ & $0.92\pm 0.00$ \\ 
climate-model-simulation-crashes (40994) & 0.07 & $0.85\pm 0.02$ & $0.74\pm 0.02$ & $0.73\pm 0.02$ \\ 
\hline 
\end{tabular}
\label{tab:auroc-results-where}
\end{table*}